\documentclass[conference]{IEEEtran}
\IEEEoverridecommandlockouts
\usepackage{cite}
\usepackage{amsmath,amssymb,amsfonts}
\usepackage{algorithmic}
\usepackage{graphicx}
\usepackage{textcomp}
\usepackage{xcolor}
\def\BibTeX{{\rm B\kern-.05em{\sc i\kern-.025em b}\kern-.08em
    T\kern-.1667em\lower.7ex\hbox{E}\kern-.125emX}}

\usepackage[inline]{enumitem}
\usepackage{hyperref}
\usepackage{color}
\usepackage{booktabs}
\hypersetup{urlcolor=cyan}

\usepackage{fancyhdr}
\begin{document}

\title{Indeterminacy in Affective Computing: Considering Meaning and Context in Data Collection Practices
{\footnotesize \thanks{This research was (partially) funded by the Hybrid Intelligence Center, a 10-year program funded by the Dutch Ministry of Education, Culture, and Science through the Netherlands Organisation for Scientific Research, \url{https://hybrid-intelligence-centre.nl}, grant number 024.004.022.}}}

\author{
 \IEEEauthorblockN{Bernd Dudzik\IEEEauthorrefmark{1}, Tiffany Matej Hrkalovic\IEEEauthorrefmark{1}\IEEEauthorrefmark{2}, Chenxu Hao\IEEEauthorrefmark{1}, Chirag Raman\IEEEauthorrefmark{1}, Masha Tsfasman\IEEEauthorrefmark{1}} 
\vspace{.5em}
\IEEEauthorblockA{
 \newline
\IEEEauthorrefmark{1}\textit{INSY, Delft University of Technology, Delft (NL), 2628XE}} 
\textit{Email: \{B.J.W.Dudzik, T.MatejHrkalovic, C.Hao-1, C.A.Raman, M.Tsafasman\}@tudelft.nl}
\IEEEauthorblockA{\IEEEauthorrefmark{2}\textit{Free University Amsterdam, Amsterdam (NL), De Boelelaan 1105, 1081 HV}}
 }

\maketitle
\thispagestyle{fancy}

\begin{abstract}
\textit{Automatic Affect Prediction (AAP)} uses computational analysis of input data such as text, speech, images, and physiological signals to predict various affective phenomena (e.g., emotions or moods). These models are typically constructed using supervised machine-learning algorithms, which rely heavily on labeled training datasets. In this position paper, we posit that all AAP training data are derived from human \textit{Affective Interpretation Processes (AIPs} resulting in a form of \textit{Affective Meaning}. Research on human affect indicates a form of complexity that is fundamental to such meaning: it can possess what we refer to here broadly as \textit{Qualities of Indeterminacy (QIs)}—encompassing \textit{Subjectivity} (meaning depends on who is making the interpretation), \textit{Uncertainty} (there is a lack of confidence regarding its correctness), \textit{Ambiguity} (meaning can contain mutually exclusive concepts) and \textit{Vagueness} (meaning can be situated at different levels in a nested hierarchy). Failing to appropriately consider QIs in AAP leads to results incapable of meaningful and reliable predictions in real-world settings. Based on this premise, our core argument is that a crucial step in adequately addressing indeterminacy in AAP is the development of data collection practices for modeling corpora that involve the systematic consideration of \begin{enumerate*} 
    \item a relevant set of QIs, and 
    \item context for the associated interpretation processes
\end{enumerate*}. 
To this end, we are \begin{enumerate*}
    \item outlining a conceptual model of AIPs and the QIs associated with the meaning these produce, in addition to a conceptual structure of relevant context, supporting understanding of its role. Finally, we use our framework for
    \item discussing examples of context-sensitivity-related challenges for addressing QIs in data collection setups. 
\end{enumerate*}  We believe our efforts can stimulate a structured discussion of both the role of aspects of indeterminacy and context in research on AAP, informing the development of better practices for data collection and analysis. 
\end{abstract}

\begin{IEEEkeywords}
Affective Computing, Ambiguity, Subjectivity, Context-awareness, User-modeling, Date Collection
\end{IEEEkeywords}

\section{Introduction}

One of the main goals in Affective Computing research is \emph{Automatic Affect Prediction (AAP)} -- providing computers with a human-like ability to detect or anticipate affective states from multi-modal sensor data \cite{Dmello2015}. Such technology has been envisioned for applications in a broad range of domains, including intelligent adaptation and personalization (e.g., in entertainment \cite{Hanjalic2006, Soleymani2013}), transformation of mental health research \cite{luxton2015artificial} and understanding consumer behavior \cite{Bouzakraoui2020}. 

Typically, \textit{Automatic Affect Prediction (AAP)} technology includes using computational analysis of input data — e.g., text, speech, images, and physiological signals — to predict various affective states (e.g., emotions or moods) in individuals or groups \cite{wang2022systematic}. These models are generally built using supervised machine learning algorithms that significantly depend on labelled training datasets \cite{wang2022systematic}. Such corpora of training data are often publicly released to the community to support modeling efforts (e.g., \cite{wang2022systematic, banziger2010introducing}) and are acknowledged as essential for the field's advances \cite{soleymani2014corpus}. 

Creating labels for these datasets can take different forms, depending on the precise prediction task. However, most of them involve humans providing some form of affective interpretation, where selected individuals are asked to either manually annotate relevant sensor data (e.g., \cite{reidsma2006annotating, navarretta2014predicting}) or self-report their impressions of how they feel (e.g., \cite{dudzik2021collecting}). The meaning of these interpretations can then serve as a label for supervision and, as such, the target of any predictions. In this discussion, we will refer to these acts of interpretation generically as \textit{Affective Interpretation Processes (AIPs)} of some specific \textit{Target Stimulus (TS)} w.r.t. some affect-related \textit{Information Goal}, while acknowledging that a plethora of Affective Science literature provides nuanced discussions of different forms of such interpretation processes\footnote{For example, under the umbrella of \textit{Emotion Causation} \cite{Moors2009} when describing the emotional interpretation of situations w.r.t. one's own feelings, or \textit{Emotion Perception} \cite{Aviezer2017} emotionally interpreting others' behavior}. This allows us to describe settings commonly captured in modelling corpora, e.g., where people provide emotional self-reports \cite{dudzik2021collecting} or label emotions for individuals depicted in media \cite{poria2018meld}. 

While the availability of ever-larger datasets containing such labels has resulted in demonstrable technological progress over the years (e.g., \cite{mollahosseini2017affectnet}), these often present an overly simplified picture of the meaning carried by the human interpretation processes \cite{dudzik2019context}. This simplification can inhibit AAP predictions from accurately capturing key aspects of human emotional functioning \cite{Dudzik2020p2, cabitza2022}. In particular, we believe that meaning, as generated by AIPs, is defined by a set of inherent \textit{Qualities of Indeterminacy (QIs)}, encompassing at least \textit{subjectivity} (i.e., its form depends heavily on who is making the interpretation); \textit{ambiguity} (what is being interpreted can have multiple plausible meanings simultaneously); \textit{uncertainty} (the correct meaning is unclear to the interpreter), and \textit{vagueness} (meaning can be situated at different levels in a nested hierarchy). 

The varying presence of these qualities in affective meaning is a fundamental part of human affective interpretations (e.g., uncertainty \cite{anderson2019}). For example, facial expressions in isolation can lead to the simultaneous perception of multiple plausible emotions \cite{cabitza2022}, and without meaningful context, interpretations can be unstable \cite{Marpaung2017}. However, these properties have not yet been systematically considered in AAP research, especially in regards to data collection \cite{dudzik2019context} (also see the motivation for \cite{dudzik2021collecting, girard2023}). Accounting for indeterminacy is crucial, as empirical evidence indicates that failing to consider indeterminacy in AAP research can result in ill-fit affective meaning in real-world phenomena, potentially leading to unreliable \cite{cabitza2022} or structurally misaligned predictions \cite{Dudzik2020p2}. 

In this article, we argue that to properly address QIs within AAP, it is essential to systematically consider how context influences the interpretation processes involved in annotation. Specifically, current AAP research not only lacks a clear conceptual understanding of the different types of indeterminacy in the affective meanings captured by labels, but also how various contextual conditions shape them. We believe that understanding how context affects QIs is crucial for AAP to have practical, real-world applications. For instance, access to additional information about antecedent events associated with some human behavioral expressions leads to more reliable affective interpretations by observers \cite{Marpaung2017}. However, it is still largely unclear in what ways such information leads to a clearer affective meaning for interpreters (e.g., which QI in our framework is affected) and under what conditions it is essential for AAPs, or merely nice to have.    

Numerous findings from empirical research highlight the prevalence of context sensitivity related to shaping the indeterminacy of affective meaning (see Greenaway et al. \cite{Greenaway2018} for a relevant overview). For example, even relatively straightforward affective interpretations of facial expressions can be highly ambiguous and dependent on the specific context in which they occur \cite{Aviezer2017}. Based on these insights, we propose that data collection efforts for AAP should focus on: \begin{enumerate*}
    \item developing methods for measuring (or otherwise capturing) relevant Qualities of Indeterminacy (QIs),  
    \item a careful consideration of possible contextual variables that could influence these QIs, and finally
    \item to systematically document these contextual variables to facilitate comparative research.
\end{enumerate*}

\section{A Conceptual Model of (Context-sensitive) Affective Interpretation Processes}

\begin{figure}[!ht]
    \caption{Conceptual Model of Affect Interpretation Processes}
    \centering
    \includegraphics[width=1\linewidth]{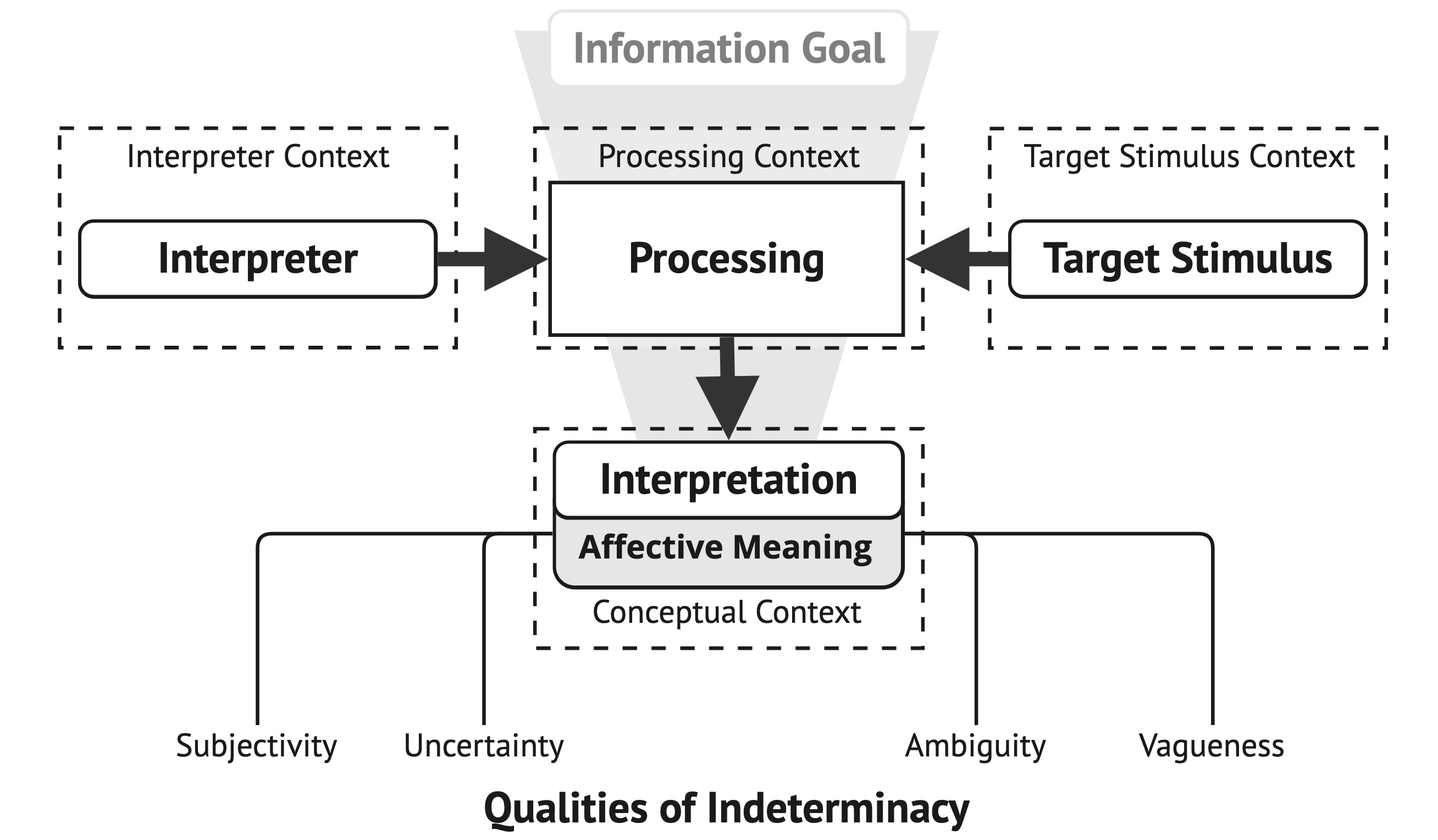}
    \label{fig:aip}
    \vspace{-2.5em}
\end{figure}

To make our arguments more precise, this section \begin{enumerate*} \item introduces a conceptual model (see \textit{Figure \ref{fig:aip}} for an overview) defining the \textit{Affective Interpretation Processes (AIPs)} in terms of a series of connected \textit{Components}, \item describes the \textit{Affective Meaning} that is produced by AIPs, focusing in particular on outlining any of its \textit{Qualities of Indeterminacy (QIs)} relevant for AAP, and \item draws on these elements to highlight \textit{Context Aspects} influencing the emergence and shape of particular QIs. \end{enumerate*} 

\subsection{Components}

\subsubsection*{\textbf{Interpreter}} refers to the human individual in which the particular Affective Interpretation Process is embedded. 

\subsubsection*{\textbf{Target Stimulus}} denotes any stimulus to be interpreted by the \textit{Interpreter} w.r.t. the current \textit{Information Goal}. Stimuli can be located externally (e.g., another person's facial expressions) or internally (e.g., a thought or a memory). However, the latter case is likely less typical in AAP settings.

\subsubsection*{\textbf{Information Goal}} describes an (implicit or explicit) affective information goal pursued in the AIP, bounding the outcome and nature of \textit{Processing} and the resulting \textit{Affective Meaning}. A property we want to highlight here is the flexibility in terms of these goals: processing could aim to interpret a \textit{Target Stimulus} regarding affective meaning for oneself (e.g., as in self-reported emotional experience \cite{cowen2017self}), relate to other person's (Emotion Perception \cite{Aviezer2017}), or even focusing on hypothetical populations (e.g., as in stereotyping \cite{hess2000}).

\subsubsection*{\textbf{Processing}} incorporates the act of interpretation where an \textit{Interpreter} processes a particular \textit{Target stimulus}, w.r.t. a specific \textit{Information Goal}.

\subsubsection*{\textbf{Interpretation}} is the outcome of the processing. Among (probably) other things, it carries a specific \textit{Affective Meaning}.

\subsection{Affective Meaning and Qualities of Indeterminacy} 

\subsubsection*{\textbf{Affective Meaning}} refers to the phenomenological content of \textit{Interpretation}, i.e., the thoughts and feelings associated with it as they are consciously experienced. It is a complex phenomenon involving a detailed mental structure that dynamically changes, depending on the specific interpretation undertaken \cite{Moors2009}. In AAP tasks, qualities of meaning are mostly captured by different categories in taxonomies or other representation schemes used for labeling (e.g., labels for Basic Emotions (\cite{ekman1992there})). We believe that certain qualities of Affective Meaning have not yet been adequately considered in AAP.

\subsubsection*{\textbf{Qualities of Indeterminacy (QIs)}} describe a series of inherent characteristics of Affective Meaning, contributing to its variability, complexity, and interpretative diversity. These qualities capture the multifaceted and fluid nature of human emotional responses and interpretations, making their meaning less straightforward, predictable, or universally agreed upon. For this article, we identify four specific qualities that we deem important for improving AAP: 

\begin{itemize}
\item{\textbf{Subjectivity}} denotes an essential quality of any form of Affective Meaning and is at the heart of many psychological theories of emotional functioning (e.g., Appraisal Theories \cite{Moors2009}). In essence, it describes the property of an interpretation process requiring the existence of an Interpreter. Furthermore, it proposes that this Interpreter's nature is intimately reflected in the structure and contents of Affective Meaning. The extent to which it does, directly impacts many AAP settings (e.g., anticipating emotional responses to media \cite{Dudzik2020p2}). 

\item{\textbf{Uncertainty}} denotes an experienced lack of confidence regarding the correctness of the Affective Meaning resulting from Processing \cite{bach2012knowing,feldmanhall2019resolving}.  Uncertainty as a property has been considered in modeling research for AAP. For example, in the form of relevant computational representations \cite{sethu2019}. However, the sense in which we understand it here is specifically a property of the experienced affective meaning itself, more closely matching a metacognitive evaluation of the certainty of one's own interpretation \cite{yeung2012metacognition}. Evidence indicates that this experience of a varying degree of confidence might be a quite common phenomenon in many task settings for AAP, e.g., interpreting faces \cite{FeldmanBarrett2019, cabitza2022}. This is likely reflected in unstable interpretations observed in settings without sufficient information associated with the Target Stimulus available to perceivers \cite{dricu2020neurocognitive, Marpaung2017}. 

\item{\textbf{Ambiguity}} in our framework refers to a quality of Affective Meaning whereby interpreting any particular Target Stimulus can contain multiple, simultaneously existing -- and even mutually exclusive -- concepts. In contrast to Uncertainty, this quality does not necessarily entail the experience of any specific degree of confidence (e.g., one can be perfectly confident in the existence of either of two equally plausible options).       

\item{\textbf{Vagueness}} is a quality denoting the relative granularity or specificity of Affective Meaning. In particular, such meaning seems to exist within a nested hierarchical structure (see, e.g., work on Emotional Granularity \cite{smidt2015brief} or theoretical frameworks aiming for abstractions in terms of descriptions \cite{cowen2017self}). As such, meaning could take shape (or at least be communicated) in terms of multiple valid ways, ranging from low granularity (e.g., a positive feeling) to high granularity of specificity (e.g., joy). 
\end{itemize}

\subsection{Context Aspects}
When talking about context, it is essential to clearly define whatever entity should be contextualized first, i.e., context should always be understood as \textit{Context-for}. For the present discussion, we are interested in Context-for any of the components of the AIPs, which we have outlined above. Consequently, we adopt the following working definition: 
\textit{\textbf{Context} is any element associated with any of AIP Component that influence properties of the \textit{Affective Meaning}}. 

Here, we briefly describe different \textit{Context Aspects} related to the presented AIP components and relevant QIs.

\subsubsection{\textbf{Interpreter Context}} refers to the interpreters' personal background that can influence the Affective Meaning of the Interpretation. In essence, this type of context is the fundamental driving force of Subjectivity, defining how the interpreter's identity is reflected in the meaning. Psychology has identified numerous aspects constituting such context \cite{Greenaway2018}). Examples range from interpreters' personal knowledge, dispositional traits, and emotional states (e.g., mood), among many others. However, there are also indications that individual differences can shape other QIs, such as Vagueness \cite{smidt2015brief}. 

\subsubsection{\textbf{Target Stimulus Context}} refers to additional information involved in processing the Target Stimulus. This information can come from various sources, such as the observable environment in which a visual stimulus is embedded in (e.g., the surrounding scene in an image). Existing findings make it plausible to assume that the QIs of Uncertainty, Vagueness, and Ambiguity could be strongly impacted by variables falling under this context aspect (\cite{hess2022infusing, dricu2020neurocognitive, Aviezer2017}).

\subsubsection{\textbf{Processing Context}} refers to any characteristics influencing Processing and resulting in changes to the Interpretation and its Affective Meaning. These could be qualitative differences in how processing takes place (e.g., conscious deliberation or largely automatic) or how it was triggered (e.g., self-initiated or prompted). We believe that Processing Context can be connected to all QIs \cite{dudzik2023valid, dreisbach2023using}.

\subsubsection{\textbf{Conceptual Context}} refers to the theoretical framework or conceptual model held by a person, shaping the content of the Affective Meaning by putting constraints on the space of possible formation and communication of meaning. Formal theoretical frameworks are often explicitly used in data collection to represent affective meaning (e.g., categories of Basic Emotions \cite{ekman1992there}). These frameworks can also more fundamentally be grounded in different languages or cultural backgrounds \cite{barrett2007language}. Following this, prior literature indicates that there is a relationship between Conceptual Context, Vagueness, Uncertainty, and Subjectivity (see, e.g., \cite{engelmann2013emotion,cowen2017self, leshin2024language}).

\section{Context Configurations as Challenges for Data Collection}
AIPs' context-sensitivity implies that contextual influences are \textit{always active} in interpretation processes, constantly shaping the resulting affective meaning. However, the strength of any particular influence may vary. We clarify this notion by introducing two different concepts when discussing contextual influences: \begin{enumerate*}
    \item a hierarchical \textit{Context Structure} of nested variables of which our Context Aspects are a high-level instantiation and
    \item a set of \textit{Context Configurations}. These configurations define the specific values that the context variables in the Context Structure take from moment to moment, ultimately shaping the outcome of any specific AIP.
\end{enumerate*}  For example, in an AIP focusing on interpreting \textit{Facial Expressions}, \textit{Gestures} could be a variable that is part of the Context Structure (Target Stimulus Context). When an actual AIP takes place, a specific facial expression will form the Target Stimulus, and the Context Configuration will contain a specific gesture. 

\subsection{Context Configurations: Phenomenon and Measurement}
Building on the notion of Context Configurations above, we outline how it relates to the design of data collection procedures. We define two types of Context Configurations that can pertain to any particular instance of an AIP:

\subsubsection{\textbf{Phenomenon Configuration}} refers to the Context Configuration as it would naturally exist when the AIP occurs. It represents the authentic, real-world conditions under which a particular Affective Meaning would emerge for it, including any associated Qualities of Indeterminacy (QIs). 

\subsubsection{\textbf{Measurement Configuration}} in contrast, refers to a specific set of arrangements as part of a data collection protocol that artificially impacts (at least some) variables of the Context Configuration to measure them. As such, it might contain values that deviate in potentially important (and sometimes unforeseen) ways from how they would occur in the Phenomenon Configuration. 

\subsection{The Influence of Measurement Configurations: Examples}
Context Configurations can significantly impact the presence and form of QIs \cite{Marpaung2017, Aviezer2017, FeldmanBarrett2019}. Therefore, it is important for researchers to be aware of the Context Structure and understand how Measurement Configurations relate or diverge from Phenomenon Configurations to make informed decisions during the design stage of the dataset collection. When establishing their Measurement Configurations, researchers must exercise caution and consider how these setups might differ from Phenomenon configurations, as such divergences can influence the QIs. Our proposed model helps identify such relations and potential divergences. Getting as close as possible to the Phenomenon  Configuration would involve minimizing any design or intervention. In practice, data collection involves trade-offs between fidelity and representativeness. To demonstrate, we offer examples of how some existing practices could represent Measurement Configurations of relevant AIP contexts and potentially impact specific QIs ($\rightarrow$).

\subsubsection{\textbf{Participant Selection sets Interpreter Context $\rightarrow$ Subjectivity}}
Selecting a group of participants with specific attributes (e.g., age \cite{cortes2021effects}, gender \cite{cortes2021effects}, or personality \cite{kafetsios2022personality}) for interpreting media content can set the Interpreter Context of an AIP in a particular way, potentially influencing the resulting Affective Meaning (Subjectivity). Conversely, by using a more homogeneous or heterogeneous pool of annotators, researchers can change the range of Interpreter Context and its effect on the Subjectivity manifested in the sample. Especially, if the sample does not align with what would constitute a relevant Phenomenon Context for where an AAP is expected to be deployed, this could lead to misaligned predictions. Therefore, the composition of annotators should be carefully considered in data collection and documented in related publications.

\subsubsection{\textbf{Questionnaire Content specifies Conceptual and Processing Context $\rightarrow$ (Uncertainty, Ambiguity, Vagueness)}} 
Selecting specific questionnaires to capture the affective meaning of an interpretation can set its boundaries. The choice of wording in a questionnaire sets the driving information goal (e.g., self-report of one's emotional experience), alters the content of the affective meaning through the type of questions used (e.g., open-ended vs. closed), and can provide a conceptual frame (e.g., answers need to be given in terms of Ekman's Basic Emotions \cite{ekman1992there}). Consequently, these factors can affect the level of vagueness, uncertainty, and ambiguity in the AIPs. Thus, the state of the questionnaires' structure should be considered during the dataset-designing process and should be properly documented.
 
\subsubsection{\textbf{Timing of Questionnaire Provision defines Processing Context $\rightarrow$ Vagueness}}
Temporal distance between the moment at which emotional self-report is asked for and the occurrence of a stimulus could also influence processing and consequently influence Affective Meaning \cite{dudzik2023valid}. For example, asking very quickly after a stimulus (e.g., part of the Measurement Configuration) could artificially prevent Processing from converging on a more specific Affective Meaning (Vagueness) than under natural circumstances (Phenomenon Configuration). Thus, the timing of label provision should be considered when selecting annotation procedures.

\section{Summary and Conclusions}    
Annotated data is essential for developing machine learning models for Automatic Affect Prediction (AAP). This article argues that any such annotations reflect Affective Meaning generated by human Affective Interpretation Processes (AIPs), which are complex and possess Qualities of Indeterminacy (QIs) that need to be accounted for in data collection for automatic predictions to be meaningful. Our core argument is that adequately addressing QIs in modeling corpora for AAP requires two crucial steps:
\begin{enumerate*}
    \item identify a set of QIs relevant for AAP and develop methods for capturing them, together with 
    \item the systematic consideration and documentation of contextual variables for the relevant interpretation processes    
\end{enumerate*}.

We are convinced that the first step is necessary, given that research in Affective Computing has not yet developed a clear vocabulary for what we have dubbed here Indeterminacy and its relationship to affective meaning. These are conceptually challenging discussions to have, likely requiring the affective computing community not only to rethink many dominant assumptions about what we (can) capture, but also how we go about modeling it. However, there is a growing awareness of the importance of elements like subjectivity in other areas of applied machine learning \cite{cabitza2023} that could prove relevant for cross-fertilization. Beyond these conceptual issues, we believe that the second step -- understanding the context-sensitivity of annotation processes -- is equally essential. Current AAP research largely lacks any clarity and data on how different varying conditions influence annotations or what impact this has on automatic predictions. As we have argued, indeterminacy captured by modeling corpora can give rise to structurally misaligned \cite{Dudzik2020p2} or unreliable predictions \cite{Marpaung2017, cabitza2022}. In our perspective, context relevant for AIPs shape the specific QIs present as part of Affective Meaning, giving rise to these challenges. Without a clearer picture of how specific changes in context shape QIs, we are largely blind to their effect or any understanding of the conditions under which AAP may operate meaningfully and reliably. Consequently, to advance AAP's real-world applicability, we advocate for prioritizing the development of data collection practices that systematically consider context aspects influencing Affective Meaning and its Indeterminacy. 

\newpage
\section*{Ethical Impact Statement}
This is a purely theoretical paper that provides an argument for adopting particular practices in specific Affective Computing Research. There are no ethical issues to discuss at the moment.

\bibliographystyle{IEEEtran}
\bibliography{EASE2024_Indeterminacy_positionPaper_arxiv}
\end{document}